\documentclass{article}
\usepackage{graphicx}
\usepackage{amssymb}
\usepackage{amstext}
\usepackage{epsfig}
\usepackage{fullpage}
\usepackage{caption}
\usepackage{subcaption}

\usepackage{url}

\usepackage{graphicx}
\usepackage{amssymb}
\usepackage{amstext}
\usepackage{epsfig}
\usepackage{caption}
\usepackage{subcaption}

\usepackage{array,multirow}

\usepackage{latexsym}
\usepackage{amsmath}

\usepackage{amsthm}
\usepackage{amsmath}
\usepackage{amsfonts}
\usepackage{amssymb,amstext}
\usepackage{appendix}
\usepackage{enumitem}

\theoremstyle{plain}
\newtheorem{theorem}{Theorem}
\newtheorem{lemma}[theorem]{Lemma}

\theoremstyle{definition}

\usepackage{algpseudocode}
\usepackage{amscd}
\usepackage{algorithm}
\usepackage{tikz}
\usetikzlibrary{arrows,decorations.markings,shapes.misc}
\tikzset{cross/.style={cross out, draw=black, minimum size=2*(#1-\pgflinewidth), inner sep=0pt, outer sep=0pt},
cross/.default={1pt}}

\newcommand{\beq}{\begin{equation}}
\newcommand{\eeq}{\end{equation}}

\newcommand{\beqa}{\begin{eqnarray}}
\newcommand{\eeqa}{\end{eqnarray}}

\newcommand{\bal}{\begin{align}}
\newcommand{\eal}{\end{align}}

\newcommand{\bsp}{\begin{equation}\begin{split}}
\newcommand{\esp}{\end{split}\end{equation}}

\newcommand{\bit}{\begin{itemize}}
\newcommand{\eit}{\end{itemize}}

\newcommand{\ben}{\begin{enumerate}}
\newcommand{\een}{\end{enumerate}}

\newcommand{\AR}{\mathbb{R}}

\newcommand{\rank}{\mathrm{rank}}

\begin{document}


\title{Top-N recommendations from  expressive recommender systems}


\author{Cyril J.~Stark \\
Massachusetts Institute of Technology\\
77 Massachusetts Avenue, 6-304\\
Cambridge, MA 02139-4307, USA \\
\texttt{cyril@mit.edu}
}

\date{November 19, 2015}

\maketitle
\begin{abstract}

Normalized nonnegative models assign probability distributions to users and random variables to items; see~\cite{AAAI16.submission}. Rating an item is regarded as sampling the random variable assigned to the item with respect to the distribution assigned to the user who rates the item. Models of that kind are highly expressive. For instance, using normalized nonnegative models we can understand users' preferences as mixtures of interpretable user stereotypes, and we can arrange properties of users and items in a hierarchical manner. These features would not be useful if the predictive power of normalized nonnegative models was poor. Thus, we analyze here the performance of normalized nonnegative models for top-N recommendation and observe that their performance matches the performance of methods like PureSVD which was introduced in~\cite{cremonesi2010performance}. We conclude that normalized nonnegative models not only provide accurate recommendations but they also deliver (for free) representations that are interpretable. We deepen the discussion of normalized nonnegative models by providing further theoretical insights. In particular, we introduce total variational distance as an \emph{operational similarity measure}, we discover scenarios where normalized nonnegative models yield \emph{unique representations} of users and items, we prove that the inference of optimal normalized nonnegative models is \emph{NP-hard} and finally, we discuss the relationship between normalized nonnegative models and nonnegative matrix factorization.

\end{abstract}

\section{Introduction}\label{sect:intro}

\begin{figure}[tbp]
\centering
\includegraphics[width=0.46\columnwidth]{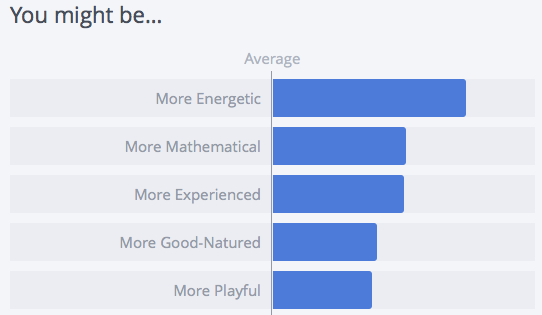}
\caption{A part of the author's personality trait as computed by OkCupid.}
\label{fig:personality.trait}
\end{figure}

Recommender systems are algorithms that are designed to help users to find interesting items. Hence, good recommender systems are able to predict which items are of interest to which users. At the same time they must be computationally tractable to handle large numbers of users and items. Consequently, recommender systems must have \emph{high predictive power} and they must be \emph{computationally tractable}. These are criteria that necessarily need to be satisfied. However, for some applications we demand more. For instance, the online dating platform OkCupid\footnote{https://www.okcupid.com/} computes for each user a `personality trait'; see figure~\ref{fig:personality.trait} for an example. These are visual representations that help users to quickly understand the personalities of other users. Thus, these personality traits can be regarded as brief sketches of other users---very much like brief characterizations of movies (e.g., through summary of the plot, list of actors, etc). Hence, visual representations of that kind complement recommendations and help a user to find what they look for.

Expressive visual representations could be drawn directly from the recommender system if the system's representation of users and items is expressive. Therefore we call a recommender system \emph{expressive} if the underlying representations of users and items are highly interpretable. We conclude that for some applications, an ideal recommender system meets the objectives \emph{high predictive power}, \emph{computational tractability} and \emph{high interpretability}.

How can we address the partially conflicting objectives `high predictive power', `computational tractability' and `high interpretability'? In~\cite{AAAI16.submission} we proposed getting inspiration from engineering and from the natural sciences. There we almost always adopt a paradigm that we might call \emph{system-state-measurement} paradigm. In~\cite{AAAI16.submission} we adopted that very same perspective for the design of recommender systems. In the context of recommender systems, the \emph{system} is that part of our mind that determines which items we like. The \emph{state} assigned to a user specifies the characteristics of that user's system, i.e., it describes that user's taste. The \emph{measurements} we perform on the system to probe the users' state (i.e., taste) are questions like ``Do you like the movie \emph{Ex Machina}?". Asking many questions of that sort (i.e., performing many measurements) we can get a refined knowledge about a user's taste (i.e., her state). In large parts of science, the system is modeled by a \emph{sample space}, the state of the system is modeled by a \emph{probability distribution} on that sample space and a measurement is modeled in terms of a random variable whose outcomes are the possible measurement outcomes. These models are called \emph{normalized nonnegative models}; see section~\ref{Sect:Model}. Normalized nonnegative models are highly interpretable because these models are identical with the highly interpretable models from the natural sciences and from engineering. 

The interpretability of normalized nonnegative models allows to \emph{draw conclusions from user and item representations}. For instance, in~\cite{AAAI16.submission} we showed how normalized nonnegative models enable us to regard users as mixtures of interpretable user stereotypes and we explained how normalized nonnegative models can be used for the computation of hierarchical orderings of properties of users and items. In sections~\ref{Sect:uniqueness}, \ref{Sect:user.user.sim.meas} and~\ref{sect:item.similarity} we introduce more features of normalized nonnegative models. In particular, we introduce an \emph{operational user-user similarity measure}, we define an \emph{operational item-item similarity measure} and we uncover scenarios where normalized nonnegative models yield \emph{unique representations} of users and items. We complement the discussion of interpretability with an empirical study (cf. section~\ref{sect:experiment}) where we test the \emph{predictive power} of normalized nonnegative models. More precisely, we evaluate the performance of these models in terms of \emph{top-$N$ recommendation} where for each user $u$, the recommender system must compile a list of $N$ items that are of interest to that user.

Apart from these practical considerations we investigate the computational complexity of the inference of optimal normalized nonnegative models (section~\ref{sect:hardness}) and we explain in what sense normalized nonnegative models are related to nonnegative matrix factorization (section~\ref{sect:reduction}).

\subsection*{Notation}\label{Sect:Notation}

For any $n \in \mathbb{N}$, $[n] = \{ 1,...,n \}$. $\AR^D_+$ denotes the set of $D$-dimensional nonnegative vectors, i.e., $\{ \vec{x} \in \AR^D |Êx_j \geq 0\}$. It contains the probability simplex $\Delta = \{ \vec{x} \in \AR^D_{+} | \sum_{j} x_{j} = 1 \}$. By $\| \vec{v} \|_p$ we denote the $l_p$ norm of a vector. For any invertible matrix $A$, $A^{-T} = (A^{-1})^T$. We will frequently refer to finite sample spaces. These are denoted by $\Omega = \{ \omega_{1}, ..., \omega_{D} \}$ where $\omega_{j}$ are elementary events. For any event $H \subseteq \Omega$ we denote its probability by $\mathbb{P}[H]$. By Kolmogorov, a random variable $\hat{E}$ on $\Omega$ with alphabet size $Z$ is a mapping $\hat{E}: \Omega \rightarrow [Z]$. We use the notation $\{ \hat{E} = z \} = \hat{E}^{-1}(z) = \{ \omega \in \Omega | \hat{E}(\omega) = z \}$. The total variational distance $\delta(\vec{p}, \vec{q})$ forms a natural distance measure between distributions $\vec{p}, \vec{q} \in \Delta$. It is defined by $\delta(\vec{p}, \vec{q}) = \frac{1}{2} \sum_{j} | p_{j} - q_{j} |$. We use $U$ to denote the number of users, $I$ to denote the number of items, $Z$ the number of different ratings (e.g., $Z = 5$ for 5-star ratings). $R_{ui}$ denotes the rating user $u$ provides for item $i$. The collection of those ratings forms the rating matrix $R \in [Z]^{U \times I}$. Recall($N$) is defined in appendix~\ref{sect:error.measrues}. We use $M \subseteq [U] \times [I]$ to mark ratings in the training set. Analogously, $T$ marks the ratings in the test set.

\section{Normalized nonnegative models}\label{Sect:Model}

By Kolmogorov, a (finite) random experiment is defined through the following triple:
\begin{itemize}
\item		A sample space $\Omega = \{ \omega_1, ..., \omega_D\}$ where $\omega_j$ are elementary events.
\item		A probability measure $\vec{p} \in \mathbb{R}^D_+$ with $\sum_j (\vec{p})_j = 1$.
\item		A random variable $\hat{E}$. This is a function $\hat{E}: \Omega \rightarrow \{ 1,...,Z \}$ for some $Z \in \mathbb{N}$.
\end{itemize}
The distribution $\vec{p}$ can be regarded as the \emph{state} of the system under consideration and the random variable $\hat{E}$ can be regarded as the description of the \emph{measurement} we perform on that system. This interpretation of distributions and random variables is ubiquitous in science and engineering. In~\cite{AAAI16.submission} we adopt the same perspective for the description of how users rate items. More specifically we denoted by $R_{ui}$ user $u$'s rating of item $i$. In case of 5-star-ratings, $R_{ui} \in \{1,..,5\} = [5]$, and more generally, $R_{ui} \in [Z]$ for some $Z \in \mathbb{N}$. We described the taste of a user $u$ in terms of a probability distribution $\vec{p}_u$ on some (unknown) sample space $\Omega = \{ \omega_1, ..., \omega_D \}$, and we regarded the process of asking user $u$ to rate item $i$ as a `measurement' we perform on the user's taste. Therefore, we modeled the question ``How do you rate item $i$?" in terms of a random variable $\hat{E}_i: \Omega \rightarrow [Z]$. The outcome $\hat{E}_i$ is the rating of item $i$.

Let $\mathbb{P}_u[ \hat{E}_{i} = z ]$ be the probability for user $u$ to rate item $i$ with value $z$. This probability can be expressed as follows. 
\begin{equation}\begin{split}\label{fwekjfsjdjkskkk}
	\mathbb{P}_u[ \hat{E}_{i} = z ]
	&=	\mathbb{P}_u[ \hat{E}_{i}^{-1}(z) ] 
	=	\sum_{j=1}^D (\vec{p}_u)_j (\vec{E}_{iz})_j 
\end{split}\end{equation}	
where $(\vec{E}_{iz})_j = 1$ if $\omega_j \in \hat{E}_{i}^{-1}(z)$, and $(\vec{E}_{iz})_j = 0$ otherwise (see~\cite{AAAI16.submission} for examples). By~\eqref{fwekjfsjdjkskkk}, the probabilities $\mathbb{P}_u[ \hat{E}_{i} = z ]$ are determined in terms of an inner product between nonnegative vectors $\vec{p}_u$ and binary vectors $\vec{E}_{iz}$ satisfying $\sum_{z=1}^Z \vec{E}_{iz} = (1,...,1)^T$. Thus, item $i$ is described by vectors $\vec{E}_{i1}, ..., \vec{E}_{iZ}$ satisfying $\sum_{z=1}^Z \vec{E}_{iz} = (1,...,1)^T$. We denote by $\mathcal{E}$ the set of allowed vectors $\bigl( \vec{E}_{i1}, ..., \vec{E}_{iZ} \bigr) \in \{0,1\}^{DZ}$, and we denote by $\Delta \subset \mathbb{R}^D_+$ the set of all probability distributions $\vec{p}_u$. The set $\Delta$ is convex. However, $\mathcal{E}$ is not. For computational reasons, we relax $\mathcal{E}$ to its convex hull $\mathcal{E}'$. 

We end up with \emph{normalized nonnegative models}: the probability distribution over ratings $z \in [Z]$ is given by $\bigl( \vec{p}_u^T \vec{E}_{iz} \bigr)_{z \in [Z]}$ for some vectors $\vec{p}_u \in \Delta$ and $\bigl( \vec{E}_{i1}, ..., \vec{E}_{iZ} \bigr) \in \mathcal{E}'$. Here, $\bigl( \vec{E}_{i1}, ..., \vec{E}_{iZ} \bigr) \in \mathcal{E}'$ if and only if
\begin{equation}\label{wdwdw3332}
	\vec{E}_{iz} \in \mathbb{R}^D_+ \text{ and } \sum_{z=1}^Z \vec{E}_{iz} = (1,...,1)^T.
\end{equation}
This class of models is slightly different from conventional probabilistic descriptions because $\mathcal{E}'$ is equal to the convex hull of the respective set from probability theory. Operationally we can think of the relaxation $\mathcal{E} \mapsto \mathcal{E}'$ as the result of a two-step procedure. First each user measures her rating for a particular item. Then, the user second-guesses that rating and randomly reassigns the measured rating to another rating. For example a user may actually like an embarrassingly stupid comedy movie. But because she is ashamed of the honest 4-star rating she decides with probability 1/2 to provide a 3-star-rating instead of the honest 4-star rating.

The previous formulation of normalized nonnegative models was \emph{categorical}, i.e., we do not need to assume any scale or linear order of the answers from $[Z]$. This feature appears to require a lot of data for training. Therefore, in section~\ref{sect:experiment}, due to the little-data problem, we regard the star ratings as approximate probabilities for liking a particular item, i.e.,
\beq\label{dekjwkj4}
	\mathbb{P}[\text{$u$ likes $i$}] \approx R_{ui}/Z.
\eeq

\subsection{Algorithm}\label{Sect:algorithm}

A natural approach for the computation of normalized nonnegative models (NNM) uses constrained alternating optimization as described in the following algorithm~\ref{Alg:Constrained.least.squares.for.NNM}. In~\cite{AAAI16.submission} we comment on the scalability of this approach. Note that all the steps in algorithm~\ref{Alg:Constrained.least.squares.for.NNM} can be parallelized and the algorithm converges to a local minimum for root-mean-squared-error on the training set. Note that we fill missing entries in the training set with zeros to counter the selection bias towards popular items. This step is dual to the assumptions the testing procedure (cf. appendix~\ref{sect:error.measrues}) relies on, and this step was also employed in~\cite{cremonesi2010performance}.

\begin{algorithm}
\caption{Constrained least squares}\label{Alg:Constrained.least.squares.for.NNM}
\begin{algorithmic}[1]
\State Fix $D$ (e.g., by cross validation).
\State	For all $u$, sample $a_{u} \in [D]$ uniformly at random and initialize $\vec{p}_{u}$ by $\vec{p}_{u} = \vec{e}_{a_{u}}$ where $(\vec{e}_i)_j = \delta_{ij}$ is a member of the canonical basis.
\State For all $(u,i) \not\in M$, set $R_{ui} = 0$ and add $(u,i)$ to $M$.
\State	For all items $i$, solve the (linearly constrained) nonnegative least squares (NNLS) problem $ \label{efhjeh}\min_{(\vec{E}_{iz})_{z \in [Z]} \in \mathcal{E}'} \sum_{u: (u,i)\in M} \bigl( \vec{E}_{iz}^T \vec{p}_u - R_{ui}/Z \bigr)^2$.
\State	For all users $u$, solve the (linearly constrained) nonnegative least squares (NNLS) problem $\min_{\vec{p}_u \in \Delta} \sum_{i: (u,i)\in M} \bigl( \vec{E}_{iz}^T \vec{p}_u - R_{ui}/Z \bigr)^2$.\label{43j5hjhfe}
\State Repeat steps 4 and 5 until a stopping criteria is satisfied (e.g., maximum number of iterations).
\end{algorithmic}
\end{algorithm}

\section{Limited interpretability of general matrix factorizations}\label{Sect:reg.and.similarity}

In the most basic matrix factorization models (see for example~\cite{koren2009matrix}) we assign vectors $\vec{x}_u \in \mathbb{R}^D$ to users $u$ and we assign vectors $\vec{y}_i \in \mathbb{R}^D$ to items $i$. These vectors are chosen such that $\vec{x}_u^T \vec{y}_i \approx R_{ui}$. To understand limitations of that approach let us assume for simplicity that the matrix factorization technique we employ is unregularized. I.e., every family of vectors $\vec{x}_u, \vec{y}_i \in \mathbb{R}^D$ provides valid descriptions of the users and items as long as $\vec{x}_u^T \vec{y}_i \approx R_{ui}$. Apart from overfitting there is another problem with unregularized factorizations of that type: for any invertible matrix $A$ we have
\begin{equation}\label{fwfegth343}
	\vec{x}_u^T \vec{y}_i \approx R_{ui} \ \Leftrightarrow \ \bigl( A^{-T}\vec{x}_u \bigr)^T  A \vec{y}_i \approx R_{ui}.
\end{equation}
Hence, there are many equivalent ways to represent users and items in terms of vectors $\vec{x}_u$ and $\vec{y}_i$, respectively. This freedom significantly \emph{limits the possibility to base any interpretation of the users' behavior on geometric properties of $\vec{x}_u$ and $\vec{y}_i$}.

For instance, we cannot use $\| \vec{x}_u - \vec{x}_{u'} \|$ as user-user similarity (important in collaborative filtering~\cite{resnick1994grouplens,sarwar2001item,deshpande2004item,o1999clustering,sarwar2002recommender}. To illustrate this point, we choose $A$ proportional to the identity matrix $I$, i.e., $A = \lambda I$ for some $\lambda \neq 0$. It follows that the variation of $\lambda$ allows to choose $\| A^{-T} \vec{x}_u - A^{-T} \vec{x}_{u'} \|$ arbitrarily because
\begin{equation}\label{dgrge43}
	\| A^{-T} \vec{x}_u - A^{-T} \vec{x}_{u'} \| = \| \vec{x}_u - \vec{x}_{u'} \| / \lambda.
\end{equation}
Observation~\eqref{dgrge43} is independent of the norm $\| . \|$ we choose because $\| \vec{v}/\lambda \| = \| \vec{v} \| / \lambda $ for all norms. Another popular user-user similarity measure is cosine similarity~\cite{resnick1994grouplens,herlocker2002empirical,mclaughlin2004collaborative,sarwar2001item}. To explain why this similarity measure is not stable under the transformations~\eqref{fwfegth343}, we assume $\vec{x}_u = (1,0)$ and $\vec{x}_{u'} = (1,\varepsilon)$ for some small scalar $\varepsilon > 0$. It follows that $\angle(\vec{x}_u, \vec{x}_{u'}) \approx 0$ where $\angle(\vec{x}_u, \vec{x}_{u'})$ denotes the angle between $\vec{x}_{u}$ and $\vec{x}_{u'}$. However, choosing $A = \mathrm{diag}(1,\lambda)$ with $\lambda \gg 1/\varepsilon$, we get $\angle\left( A^{-T}\vec{x}_u, A^{-T}\vec{x}_{u'} \right) \approx \pi / 2$. Choosing $A$ more generally, we can squeeze and stretch the vectors $\{ \vec{x}_u \}_{u \in [U]}$ in arbitrary directions. 

Obviously, regularizing the matrix factorization by penalizing the norms of the vectors $\vec{x}_u$ and $\vec{y}_i$ (e.g., Tikhonov regularization) helps to restrict our freedom~\eqref{fwfegth343} to rescale user and item vectors. However, even if the solution of problems of the sort
\[
	\min \ \Bigl( \sum_{u,i} \bigl( \vec{x}_u^T \vec{y}_i - R_{ui}  \bigr)^2 \Bigr) + \mu \Bigl( \sum_u \| \vec{y}_u  \|^2 \Bigr) + \mu \Bigl( \sum_i \| \vec{y}_i \|^2 \Bigr)
\]
is unique we still do not know how to interpret the vectors $\vec{x}_u$ and $\vec{y}_i$. We do not even know whether $\| \vec{x}_u - \vec{x}_{u'} \|$ accurately reflects the actual similarity of users $u$ and $u'$. In fact it is easy to come up with two regularization procedures that both guarantee uniqueness but lead to disagreeing distances between vectors we assign to users. To summarize we note that 
\begin{itemize}
\item different regularization schemes may or may not lead to unique determination of geometric quantities like $\| \vec{x}_{u} - \vec{x}_{u'} \|$ that we wish to interpret operationally.
\item	But even among the regularization schemes that uniquely determine geometric quantities like $\| \vec{x}_{u} - \vec{x}_{u'} \|$, a change of the regularization leads to a change of the values $\| \vec{x}_{u} - \vec{x}_{u'} \|$.
\end{itemize}
Hence, regularization, geometric interpretability and similarity are \emph{intimately related to each other}. Nonnegative matrix factorization (NMF) has become popular because of the interpretability of the user and item vectors. However, NMF without norm penalization (Tikhonov) suffers from the same issues mentioned before as the transformation $A$ from our examples maps vectors from $\mathbb{R}^D_+$ back into $\mathbb{R}^D_+$. NMF is particularly tricky as the freedom described by $A$ cannot only be used to alter similarity measures, it can also be used to change our interpretation of the user and item vectors.  For instance, a positive matrix $A= \mathrm{diag}(\lambda_1,...,\lambda_D)$ can be used to arbitrarily shrink or stretch vectors $\vec{x}_u$ in all directions. Consequently, in one NMF we might have $\vec{x}_u = (1, \varepsilon, ..., \varepsilon)^T$ leading to the conclusion that user $u$ is very well described by the feature associated with $(1,0,...,0)^T$. On the other hand, for $A = \mathrm{diag}(1/\epsilon, 1,...,1,\epsilon)$ we have that $A^{-T} \vec{x}_u = (\varepsilon, ..., \varepsilon,1)^T$ leading to the conclusion that user $u$ is accurately described by the feature represented by $(0,...,0,1)^T$.

\section{Uniqueness}\label{Sect:uniqueness}

We claim that ambiguities of the form~\eqref{fwfegth343} are addressed in normalized nonnegative models. Intuitively, one expects the user vectors $\vec{p}_u$ and the item vectors $\vec{E}_{iz}$ to be approximately uniquely defined if the entries in the rating matrix $R$ push these vectors towards the boundary of the cone $\mathbb{R}^D_+$. That is because in these situations, the set of allowed states $\Delta$ and the set of allowed measurements $\mathcal{E}'$ leave no room to wiggle the user and item vectors as in~\eqref{fwfegth343}. To confirm this intuition we consider an toy example that we can analyze rigorously. Assume that $D = Z$ and assume that the users do not just provide $R$. Instead, for each $u,i,z$, user $u$ provides us with an accurate estimate of the probability to rate $i$ with $z$ `stars'. Moreover, we assume that for each $i,z$ there exists at least one user $u_{iz}$ who rates $i$ with $z$ `stars'. We claim that datasets of that kind uniquely determine the underlying normalized nonnegative model.

To prove this claim we first observe that by Cauchy-Schwarz, 
\beq\label{fwefl4rkl}
	1 = \mathbb{P}_{u_{iz}}[\hat{E}_i = z] = \vec{p}_{u_{iz}}^T \vec{E}_{iz} \leq \| \vec{p}_u \|_2 \| \vec{E}_{iz} \|_2 \leq \| \vec{E}_{iz} \|_2
\eeq
for all $i,z$ because $\| \vec{p} \|_2 \leq 1$ for every probability distribution. For any $\vec{E}_{iz}, \vec{E}_{iz'} \in \AR^D_+$ we have that $\vec{E}_{iz}^T \vec{E}_{iz'} \geq 0$. Therefore, using $\| \vec{v} \|_2^2 = \vec{v}^T \vec{v}$,
\beq\label{fwjhjkh32}
	D = \| (1,...,1)^T \|_2^2 = \| \sum_z \vec{E}_{iz} \|_2^2 \geq \sum_z \vec{E}_{iz}^T \vec{E}_{iz} \geq Z = D.
\eeq
where we used~\eqref{fwefl4rkl} in the last inequality. Equation~\eqref{fwjhjkh32} can only be satisfied if $\vec{E}_{iz}^T \vec{E}_{iz'} = \delta_{zz'}$. This in turn can only be satisfied if each of the \emph{nonnegative} vectors $\vec{E}_{iz}$ is equal to an element of the orthonormal basis defining $\AR^D_+$. It follows that $\vec{p}_{u_{iz}} = \vec{E}_{iz}$ because this is the only possibility to satisfy $1 = \mathbb{P}_{u_{iz}}[\hat{E}_i = z]$. This concludes the proof of the claim.

We note that this sequence of arguments crucially depends on both the \emph{conic structure} (i.e., $\AR^D_+$) of NNMs, as well as on the \emph{normalization conditions} of normalized nonnegative models.

\section{Operational user similarity}\label{Sect:user.user.sim.meas}

Assume we describe users and items in terms of a normalized nonnegative model. In this section we are going to motivate the use of the total variational distance $\delta(\vec{p}_x, \vec{p}_{x'}) = \frac{1}{2} \sum_{j=1}^D | (\vec{p}_x)_j - (\vec{p}_{x'})_j|$ as user-user similarity measure by a game which provides an operational interpretation of $\delta(\vec{p}_x, \vec{p}_{x'})$. Imagine two users $u=1$, $u=2$ and an item $i$. Assume you know the representations $\vec{p}_1, \vec{p}_2 \in \Delta$ of the users' tastes and assume you know the description $(\vec{E}_{i1}, ..., \vec{E}_{iZ}) \in \mathcal{E}'$ of item $i$. Now 
\begin{enumerate}
\item a referee flips an unbiased coin to select a user $\hat{u} \in \{1,2\}$. 
\item	The referee asks user $\hat{u}$ to rate item $i$. We denote the value of $\hat{u}$'s rating by $r$.
\item Then, the referee hands you a note specifying $r$ but not $\hat{u}$. 
\item Your objective is to guess $\hat{u}$.
\end{enumerate}
We can compute all the probabilities $\mathbb{P}_u[\hat{E}_i = r]$ for user $u \in \{1,2\}$ to rate $i$ with $r$ because we know the descriptions $\vec{p}_1, \vec{p}_2$ of the users and the description $(\vec{E}_{i1}, ..., \vec{E}_{iZ})$ of the item. If $\mathbb{P}_1[\hat{E}_i = r] > \mathbb{P}_2[\hat{E}_i = r]$ then we better guess that user $1$ provided the rating. Otherwise, we guess that user $2$ provided the rating. Set $\mathcal{Z} = \bigl\{ z \in [Z] \ \bigl| \ \mathbb{P}_1[\hat{E}_i = z] > \mathbb{P}_2[\hat{E}_i = z] \ \bigr\}.$ Hence, we guess $\hat{u} = 1$ if and only if $r \in \mathcal{Z}$. This (optimal) strategy succeeds with probability $p_{\mathrm{success}} = \frac{1}{2} \mathbb{P}_1[\hat{E}_i \in \mathcal{Z}] + \frac{1}{2} \mathbb{P}_2[\hat{E}_i \not\in \mathcal{Z}]$ (recall that the coin is unbiased). By complementarity of the events $\{ \hat{E}_i \in \mathcal{Z} \}$ and $\{ \hat{E}_i \not\in \mathcal{Z} \}$,
\begin{equation}\begin{split}\label{fwekjwwdfsjdjkskkk}
	p_{\mathrm{success}} 
	=		\frac{1}{2} \Bigl(  1 + \mathbb{P}_1[\hat{E}_i \in \mathcal{Z}]- \mathbb{P}_2[\hat{E}_i \in \mathcal{Z}] \Bigr)
	\leq		\frac{1}{2} \Bigl(  1 +  \delta(\vec{p}_1, \vec{p}_2) \Bigr)
\end{split}\end{equation}	
where $\delta(\vec{p}_1, \vec{p}_2)$ denotes the \emph{total variational distance}, i.e.,
\begin{equation}\label{fewknk4}
	\delta(\vec{p}_1, \vec{p}_2) = \max_{\Omega_0 \subseteq \Omega} | \mathbb{P}_1[ \Omega_0 ] - \mathbb{P}_2[ \Omega_0 ] | = \frac{1}{2} \| \vec{p}_1 - \vec{p}_2 \|_1
\end{equation}	
(see~\cite{cover2012elements}). We conclude that $\delta(\vec{p}_1, \vec{p}_2)$ yields an upper bound on our success probability which is independent of the item $i$. More importantly, however, this upper bound is tight, meaning that there exists a hypothetical item $i^*$ that leads to a success probability $p_{\mathrm{success}}^*$ satisfying
\begin{equation}\label{fewknk4wdw}
	p_{\mathrm{success}}^* = \frac{1}{2} \Bigl(  1 +  \delta(\vec{p}_1, \vec{p}_2) \Bigr).
\end{equation}	
The item $i^*$ is any item satisfying $\{ \hat{E}_i \in \mathcal{Z} \} = \Omega^*$ where $\Omega^*$ is the maximizer from~\eqref{fewknk4}. Identity~\eqref{fewknk4wdw} captures the operational meaning of the total variational distance: $\delta(\vec{p}_1, \vec{p}_2)$ determines via~\eqref{fewknk4wdw} the maximal success probability for distinguishing the users $u_1$ and $u_2$. \emph{This motivates using $1 - \delta(\vec{p}_1, \vec{p}_2)$ as similarity measure because $\delta(\vec{p}_1, \vec{p}_2)$ is small if and only if users $u_1, u_2$ are difficult to distinguish.} In appendix~\ref{sect:alternative.interpretation.oftot.var.dist} we remind the reader of an alternative interpretation of $\delta(\vec{p}_1,\vec{p}_2)$. We expect $\delta(\vec{p}_1,\vec{p}_2)$ to be particularly useful to create matches on dating websites.

\section{Operational item similarity}\label{sect:item.similarity}

To derive an operational item-item similarity measure we consider a game similar to the game from section~\ref{Sect:user.user.sim.meas}. More precisely, assume you know the representations $(\vec{E}_{11}, ..., \vec{E}_{1Z})$, $(\vec{E}_{21}, ..., \vec{E}_{2Z})$ of two items $i=1,2$ and assume you know the representation $\vec{p}$ of a user $u$. The game proceeds as follows.

\begin{enumerate}
\item A referee flips an unbiased coin to select an item $\hat{i} \in \{1,2\}$. 
\item	The referee secretly asks user $u$ to rate item $\hat{i}$. We denote the value of that rating by $r$.
\item The referee hands you a note specifying $r$ but not $\hat{i}$. 
\item Your objective is to guess $\hat{i} \in \{ 1,2 \}$.
\end{enumerate}

Knowing $\vec{p}$ and both item representations, we can compute the two distributions $\vec{q} := (\mathbb{P}[ r | \hat{i} = 1 ])_{r \in [Z]}$ and $\vec{q}' := (\mathbb{P}[ r | \hat{i} = 2 ])_{r \in [Z]}$. Hence, we can rephrase the considered game: we get the value $r$ sampled from either $\vec{q}$ or $\vec{q}'$. Our task is to guess which of the two alternatives applies. According to section~\ref{Sect:user.user.sim.meas}, $p_{\mathrm{success}} = \frac{1}{2}( 1+ \frac{1}{2}\| \vec{q} - \vec{q}' \|_1)$ (`=' because this time we can choose the strategy freely). What is the optimal success probability when varying $\vec{p}$? Note that
\begin{equation}\begin{split}\nonumber
	&\max_{\vec{p} \in \Delta} 	\| \vec{q} - \vec{q}' \|_1 \ \text{s.t.} \Bigl\{ \ q_z = \vec{p}^T\vec{E}_{1z}, q'_z = \vec{p}^T\vec{E}_{2z} \; \forall z \in [Z] \Bigr\} \\
	=
	&\max_{j \in [D]}  \| \vec{q} - \vec{q}' \|_1 \ \text{s.t.} \Bigl\{ \ q_z = (\vec{E}_{1z})_j, q'_z = (\vec{E}_{2z})_j \; \forall z \in [Z] \Bigr\}.
\end{split}\end{equation}
Here, we used that maximization of $\| . \|_1$ is a LP and therefore achieved at an extremal point of $\Delta$. Therefore, we end up with the item-item similarity measure $1 - \delta(i_1,i_2)$ where
\begin{equation}\begin{split}\label{fwkj4h5jh}
	\delta(i_1,i_2) 
	&= \frac{1}{2} \max_{j \in [D]}  \bigl\|  ( E_1 - E_2 )_{j,:} \bigr\|_1
\end{split}\end{equation}
with $E_i := ( \vec{E}_{i1}, ...,\vec{E}_{iZ})$; $i = 1,2$ and $(E_1-E_2)_{j,:}$ denotes the $j$-th row of $E_1-E_2$. Using~\eqref{fwkj4h5jh}, we show in appendix~\ref{sect:application.of.item.item.similarity} how \emph{all} applications from~\cite{AAAI16.submission} can be lifted to situations where the available data does not specify any item tags.

\section{Empirical study}\label{sect:experiment}

Running Algorithm~\ref{Alg:Constrained.least.squares.for.NNM}, we evaluate the performance of normalized nonnegative models on the MovieLens 1M dataset from~\cite{miller2003movielens}. This allows us to compare our results with the results obtained in the literature; e.g.,~\cite{cremonesi2010performance}. We are interested in the part of the MovieLens dataset which specifies a long list of triples $(u,i,R_{ui})$ where $u$ is a user, $i$ is an item (i.e., a movie) and $R_{ui} \in [5]$ is the 5-star-rating of $i$ by $u$. To define the training data and the test data we proceed exactly as in~\cite{cremonesi2010performance}, i.e., we randomly sample 1.4\% of the provided movie ratings. These ratings form the test set $T$. The remaining entries form the training set $M$. In appendices~\ref{sect:error.measrues} and~\ref{sect:gergregkj} we remind the reader of the definition of recall at $N$, and we sketch the evaluation methodology~\cite{cremonesi2010performance} which we adopt in the following.

\begin{figure}
\centering
\begin{subfigure}{.5\textwidth}
  \centering
  \includegraphics[width=1\linewidth]{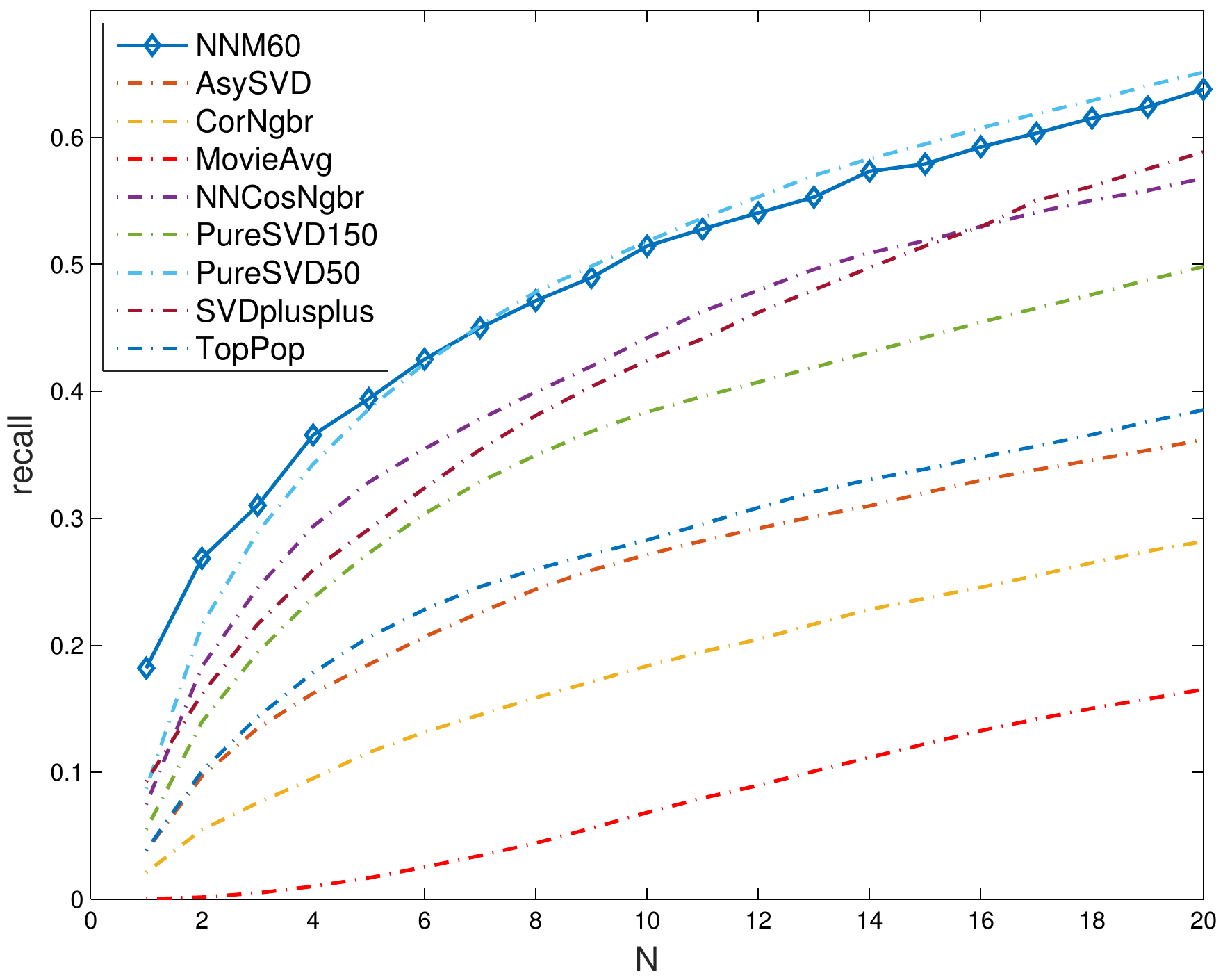}
\end{subfigure}%
\begin{subfigure}{.5\textwidth}
  \centering
  \includegraphics[width=1\linewidth]{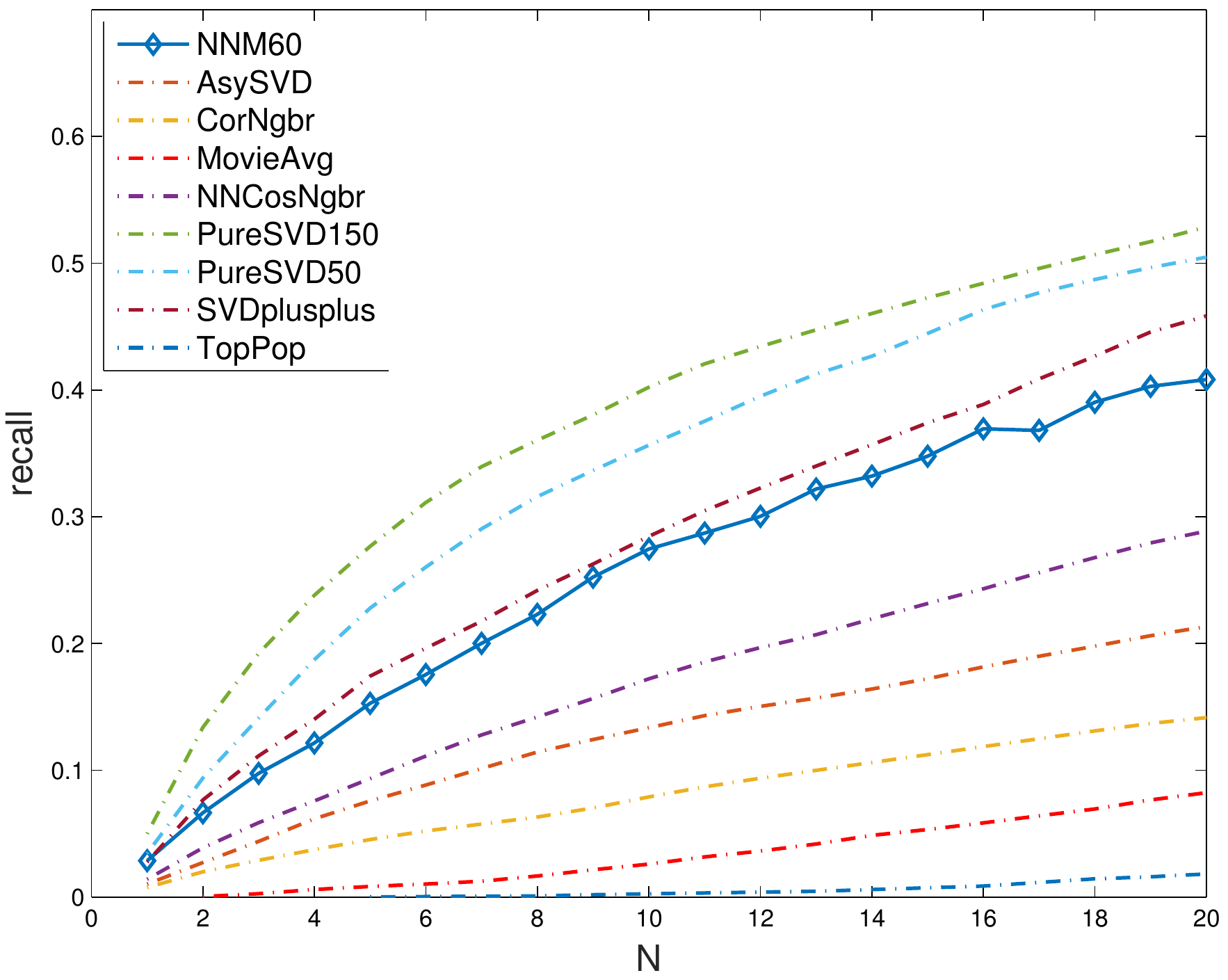}
\end{subfigure}
\caption{\emph{Left:} Recall at $N$; all items. \emph{Right:} Recall at $N$; items from long tail.}
\label{fig:recall.60D.NNM}
\end{figure}

\begin{figure}
\centering
\begin{subfigure}{.5\textwidth}
  \centering
  \includegraphics[width=1\linewidth]{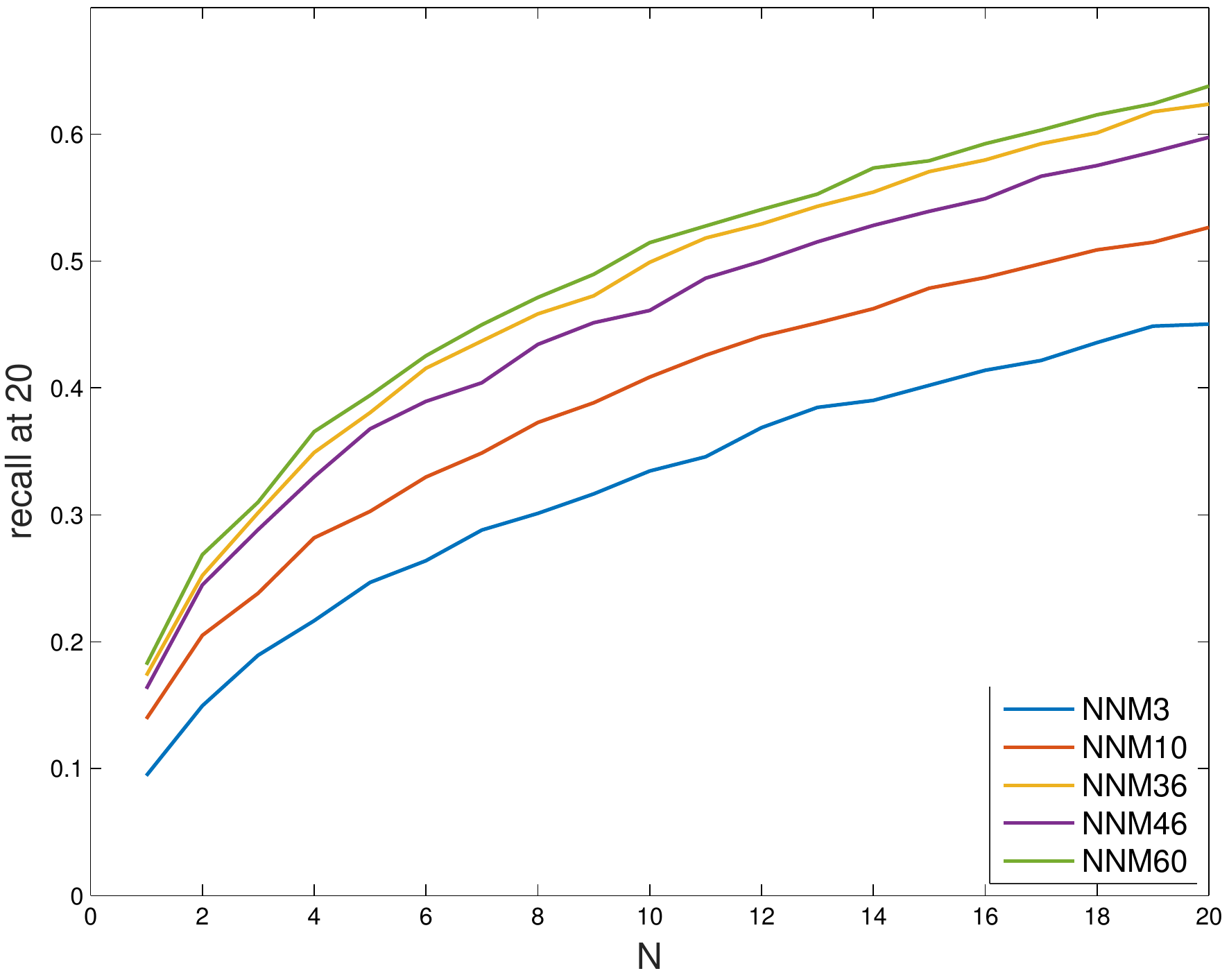}
\end{subfigure}%
\begin{subfigure}{.5\textwidth}
  \centering
  \includegraphics[width=1\linewidth]{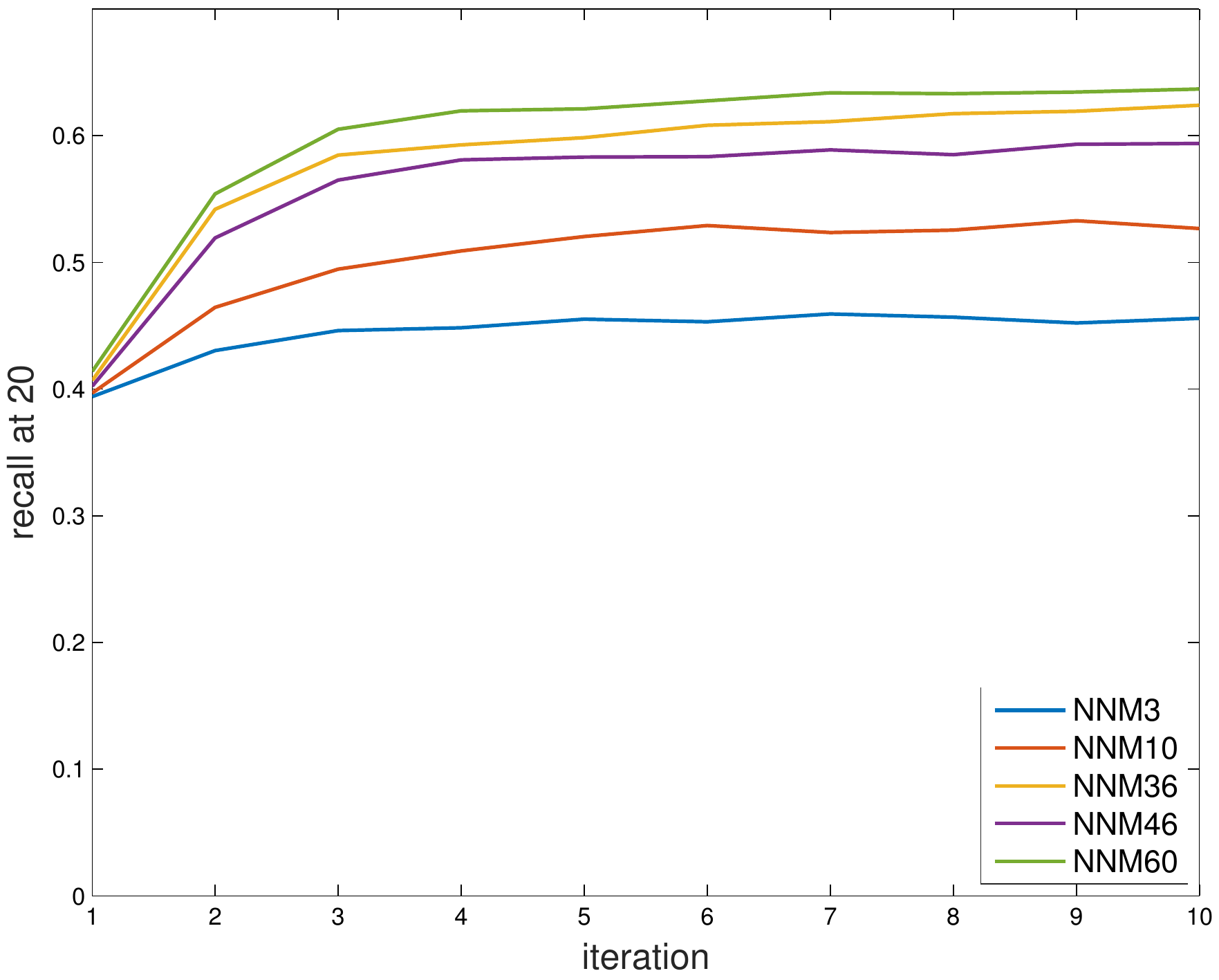}
\end{subfigure}
\caption{\emph{Left:} Recall at $N$ (all items); comparison of different model dimensions. \emph{Right:} Recall at 20 (all items) as function of iteration.}
\label{fig:recall.as.fn.of.dim.and.iteration.NNM}
\end{figure}

Figure~\ref{fig:recall.60D.NNM} (left) compares normalized nonnegative models (computed using 10 iterations) with some common recommender systems (see~\cite{cremonesi2010performance} for details). We observe that normalized nonnegative models perform particularly well for low $N$. This might be of interest in applications because we do not want to present long lists of recommendations to users. Figure~\ref{fig:recall.60D.NNM} (right) also compares normalized nonnegative models with other recommender systems. This time, however, we only take into account items from the long tail for the evaluation of recall (cf.~appendix~\ref{sect:gergregkj}). Finally, figure~\ref{fig:recall.as.fn.of.dim.and.iteration.NNM} displays recall at 20 as function of both the dimension of the model or as function of the iteration (in Algorithm~\ref{Alg:Constrained.least.squares.for.NNM}). All of these results were computed on a desktop computer (4 cores) running Matlab.

\section{Computational complexity}\label{sect:hardness}

We consider the categorical setting described in equation~\eqref{fwekjfsjdjkskkk} and imagine that (instead of star ratings) the users provide estimates for $\mathbb{P}_u[ \hat{E}_{i} = z ]$. What is the computational complexity of finding the lowest-dimensional normalized nonnegative model for the data $\mathbb{P}_u[ \hat{E}_{i} = z ]$? In other words, what is the computational complexity of problem \emph{MinDim} defined as follows.

\emph{MinDim}. Find the minimal dimension $D$ such that there exist $\vec{p}_{u} \in \Delta$ and $\vec{E}_{iz} \in \mathcal{E}'$ with the property $\vec{p}_{u}^T \vec{E}_{iz} = \mathbb{P}_u[ \hat{E}_{i} = z ]$ for all $u,i,z$.

In appendix~\ref{sect:proof.thm.NP.hardness} we prove the following theorem~\ref{Thm:NP} by showing that the natural decision version of \emph{MinDim} is \emph{NP}-hard.

\begin{theorem}\label{Thm:NP.mindim}
	The decision problem \emph{MinDim} is \emph{NP}-hard.
\end{theorem}

\section{Relation to nonnegative matrix factorization}\label{sect:reduction}

Assume we are dropping all the normalization constraints on $\vec{p}_{u}$ and $\vec{E}_{iz}$. I.e., instead of searching for a NNM with the property $\mathbb{P}_u[ \hat{E}_{i} = z ] = \vec{p}_u^T \vec{E}_{iz}$ we search for \emph{arbitrary} nonnegative vectors $\vec{a}_u , \vec{b}_{iz} \in \AR^D_+$ satisfying $\mathbb{P}_u[ \hat{E}_{i} = z ] = \vec{a}_u^T \vec{b}_{iz}$. The following Lemma~\ref{Lemma:fjekfjkejf} characterizes the relationship between models $\vec{a}_u$, $\vec{b}_{iz}$ for $\mathbb{P}_u[ \hat{E}_{i} = z ]$ on the one hand and NNMs for $\mathbb{P}_u[ \hat{E}_{i} = z ]$ on the other hand.

\begin{lemma}\label{Lemma:fjekfjkejf}
	Let $A \in \AR^{D \times U}_+$, $B \in \AR^{D \times IZ}_+$ be arbitrary nonnegative matrices with the properties
	\beq\label{kejrgkrejgj}
		\bigl( \mathbb{P}_u[ \hat{E}_{i} = z ] \bigr)_{u \in [U];i \in [I],z \in [Z]} = A^T B
	\eeq
	and $D = \rank(A) = \rank(B)$. We denote by $\vec{a}_{u}$ the columns of $A$ and by $\vec{b}_{iz}$ the columns of $B$. Then, there exists an invertible matrix $T$ such that
	\[
		\vec{a}'_u := T \vec{a}_u, \ \ \vec{b}'_{iz} := T^{-1} \vec{b}_{iz}
	\]
	is a normalized nonnegative model satisfying $\mathbb{P}_u[ \hat{E}_{i} = z ] =      \vec{a}_u'^T \vec{b}'_{iz}$ for all $u,i,z$.
\end{lemma}

Factorizations~\eqref{kejrgkrejgj} are of great importance in many disciplines; factorizations of that type are called nonnegative matrix factorizations (NMF;~\cite{lee1999learning}). Note, however, that Lemma~\ref{Lemma:fjekfjkejf} does not imply a general equivalence between NMF and NNM because to establish the transformation from NMF to NNMs and vice versa we considered the noiseless scenario and we assumed that $D = \rank(A) = \rank(B)$ (see Lemma~\ref{Lemma:fjekfjkejf}). We prove Lemma~\ref{Lemma:fjekfjkejf} in appendix~\ref{sect:proof.lemma.NMF.vs.NNM}.

\section{Related work}

We mentioned related work when discussing similarity measures, time complexity and relation to NMF. Therefore, we describe here relations between NNMs and other models for item recommendation. We start with \emph{Probabilistic matrix factorization} (PMF, see~\cite{mnih2007probabilistic,salakhutdinov2008bayesian}) which forms intriguing family of models related to NNMs. In PMFs, the rating $R_{ui}$ of user $u$ for item $i$ is regarded as Gaussian random variable. Mean and variance of $R_{ui}$ are modeled in terms of $\vec{U}_u^T \vec{V}_i$ and $\sigma$, respectively. Here, $\vec{U}_u, \vec{V}_i$ are low-dimensional vectors assigned to users and items. This structure is reminiscent of~\eqref{fwekjfsjdjkskkk} and~\eqref{fwfefeg}. However, to apply PMF we need to assume that $R_{ui}$ is Gaussian. In contrast, to apply NNMs, we do not need to assume anything about the distribution of $R_{ui}$ and $R_{ui}$ can be treated as \emph{categorical} random variable. The interpretability of PMFs is high because in principle, they allow for the computation of hierarchical orderings of properties of users and items (through the game introduced in~\cite{AAAI16.submission}). But due to the infinite-dimensional nature of PMFs (i.e., $| \Omega | = \infty$), the behavior of users cannot be interpreted easily through stereotypes; see~\cite{AAAI16.submission}.

In applications like top-$N$ recommendation \cite{herlocker2004evaluating} we are not primarily interested in ratings of items but we are interested only in the co-occurrence of pairs $(u,i)$ in measured data. The occurrence of a pair $(u,i)$ can be interpreted, for instance, as ``$u$ likes movie $i$", ``$u$ attends concert $i$", etc. When using the graphical \emph{aspect models} \cite{hofmann1999latent,hofmann1999probabilistic,blei2003latent,blei2004hierarchical} to describe these settings we regard $(u,i)$ as a two-dimensional random variable whose distribution has the form
\beq\label{fwfefeg}
	\mathbb{P}[u,i] = \sum_{k=1}^K \; \mathbb{P}[u|k] \; \mathbb{P}[i|k].
\eeq
Hence, aspect models are described in terms of a latent variable $k \in [K]$ and we observe that $u$ and $i$ are independent when conditioned on $k$. By~\eqref{fwfefeg}, $\mathbb{P}[u,i]$ can be regarded as inner product between two distributions (i.e., $( \mathbb{P}[u|k] )_{k \in [K]}$ and $( \mathbb{P}[i|k] )_{k \in [K]}$) on some sample space $\Omega = \{ \omega_1, ..., \omega_K \}$. In that sense, aspect models~\eqref{fwfefeg} are related to normalized nonnegative models~\eqref{fwekjfsjdjkskkk}. The difference lies in the \emph{different interpretation} of the vectors whose inner product we compute. In case of normalized nonnegative models we compute the inner product between a distribution (describing the user) and a convexly relaxed indicator function (describing one particular rating $z$ of the item). In case of aspect models we compute the inner product between two distributions---the first distribution describes the user and the second distribution describes the item. In practice, this distinction reveals itself (for example) when we try to use aspect models to model 5-star ratings (an instance of ratings with multiple outcomes). Describing such ratings with normalized nonnegative models is straightforward (cf.~section~\ref{Sect:Model}). On the other hand, describing ratings with multiple outcomes with aspect models is more involved because we need to decide on a particular graphical model (cf. section~2.3 in~\cite{hofmann1999latent}) and this makes the practical application of aspect models more challenging. Moreover, to the best of our knowledge, it has not yet been discussed carefully in what precise sense aspect models give rise to operational similarity measures like the total variational distance.

Regarding the empirical evaluation of top-$N$ recommendation we would like to point out the closely related works~\cite{cremonesi2010performance} and~\cite{barbieri2011analysis}. Both of these beautiful works show the disagreement between recall and precision on the one hand and RMSE on the other hand. The paper~\cite{cremonesi2010performance} introduces PureSVD for top-$N$ recommendation. Moreover, it proposes a construction of the test set that we adopt here, namely, the exclusion of most popular items to counteract the selection bias in the MovieLens dataset. The paper~\cite{barbieri2011analysis} analyzes the performance of major probabilistic models for top-$N$ recommendation.

In section~\ref{sect:reduction} we have seen how normalized nonnegative models and nonnegative matrix factorization (NMF) are related to each other. NMF plays an important role in recommendation in general; see~\cite{ma2011recommender}. Of course, probabilistic models can be regarded as a conveniently regularized NMF but that perspective disregards the operational interpretation of the columns of the nonnegative matrices $A$, $B$ that constitute the NMF $A^T B$. This interpretation is important and the choice of regularization of an NMF affects the predictive performance of the studied recommender system.

\section{Conclusion}\label{sect:conclusions}

We evaluated normalized nonnegative models in the context of item recommendation and we extended our understanding of these models; both from the practical and theoretical perspective. We deepened the practical understanding of normalized nonnegative models by studying their performance in top-$N$ recommendation and by introducing user-user and item-item similarity measures which can be interpreted operationally in terms of the distinguishability of users and the distinguishability of items. On the theoretical side we extended our understanding of normalized nonnegative models \emph{by} showing how the regularization scheme defining normalized nonnegative models can enforce unique user and item representations, \emph{by} proving that the inference of optimal normalized nonnegative models is \emph{NP}-hard and \emph{by} explaining how normalized nonnegative models are related to nonnegative matrix factorizations.

\section{Acknowledgments}

I thank Robin Kothari, Patrick Pletscher and Sharon Wulff for interesting and fruitful discussions. I thank the authors of~\cite{cremonesi2010performance,barbieri2011analysis} for providing the source files for the figures in~\cite{cremonesi2010performance,barbieri2011analysis}. I acknowledge funding by the ARO grant Contract Number W911NF-12-0486. This work is preprint MIT-CTP/4738.

\appendix

\section{Application of item-item similarity: characterization of stereotypes and hierarchical orderings}\label{sect:application.of.item.item.similarity}

In~\cite{AAAI16.submission} we explained how normalized nonnegative models can be used to succinctly describe user and items in terms of interpretable tags of items. This requires that the available data specifies tags for each item. What if no tags are available?  For these circumstances we suggest using $k$-medoids to classify the items with respect to the item-item similarity measure~\eqref{fwkj4h5jh}. Here, $k = G$, i.e., $k$ equals the number of (effective) tags we want to use. Running $k$-medoids thus assigns tags $g \in [G]$ to items. Using these tags we can proceed as in sections~5.1 and~6 from~\cite{AAAI16.submission}. 

The tags we compute by $k$-medoids do not, however, come along with intuitive names like \emph{comedy}, \emph{drama} or \emph{romance}. Hence, to intuitively understand the effective tags we propose selecting popular representative items for each of the tags (e.g., movies the user knows already). The computed tags can then be described to users in terms of the representative items. For example, in movie recommendation where tags specify genres, $\text{effective genre}_3 \sim \{ \text{movie}_9, \mathrm{movie}_{17}, \text{movie}_{97}\}.$

\section{Alternative interpretation of $\delta(\vec{p}_1,\vec{p}_2)$}\label{sect:alternative.interpretation.oftot.var.dist}

In collaborative filtering we oftentimes recommend an item $i$ to user $u_2$ if $u_1$ rated and liked item $i$, and if $u_1$ and $u_2$ are similar users. Hence, the probability that users $u_1$ and $u_2$ provide different ratings for $i$ is crucial for us. This probability can be lower bounded as follows.

\begin{equation}\begin{split}\label{fwe2wdfsjdjkskkk}
	&\mathbb{P}[R_{u_1i} \neq R_{u_2i}]
	=	 1 - \mathbb{P}[R_{u_1i}= R_{u_2i}]
	=	1 -  \sum_{z} \mathbb{P}[R_{u_1i} = z, R_{u_2i} = z]\\
	&=	1 -  \sum_{z} \mathbb{P}_1[\hat{E}_i = z] \mathbb{P}_2[\hat{E}_i = z]		
	\geq	1 -  \sum_{z} \min\{ \mathbb{P}_1[\hat{E}_i = z], \mathbb{P}_2[\hat{E}_i = z] \}\\	
	&=	\delta(\vec{p}_1, \vec{p}_2).
\end{split}\end{equation}

\section{Error measures}\label{sect:error.measrues}

When evaluating the performance of recommender systems, the choice for quantifying the prediction error crucially affects the evaluation; the observation that a system $A$ performs better than a system $B$ oftentimes changes if we change the particular way we quantify the prediction error. 

\emph{Root-mean-squared-error} (RMSE) and \emph{mean-average-error} (MAE) are popular choices for measuring the prediction error. If we denote by $\hat{R}_{ui}$ the rating predicted by our recommendation system, then $\mathrm{RMSE} = \bigl( \sum_{(ui) \in T} (R_{ui} - \hat{R}_{ui})^2 \bigr) / |T|$ and $\mathrm{MAE} = \bigl( \sum_{(ui) \in T} \bigl| R_{ui} - \hat{R}_{ui} \bigr| \bigr) / |T|$. It follows that $\mathrm{RMSE}$ and $\mathrm{MAE}$ can be regarded as $l_{2}$-distance and $l_{1}$-distance between prediction and ground truth. Being an instance of an $l_{2}$-type distance, RMSE is sensitive to outliers in $( \hat{R}_{ui} )_{ui \in T}$ whereas MAE is not. In practice, however, it is oftentimes not of immediate interest to predict actual ratings. Instead we are interested in presenting to the user a short list of items that are of interest to that user. This short list of recommendations is the \emph{top-$N$ recommendation}. \emph{Precision} and \emph{recall} are error measures designed to measure the usefulness of these top-$N$ recommendations that are computed by recommender systems. Following~\cite{cremonesi2010performance}, we define recall and precision through the following procedure. Fix $N$. Then, for each triple $(u,i,R_{ui}) \in T$ \emph{satisfying} $R_{ui} = 5$, 
\begin{enumerate}
\item sample 1000 items not rated by user $u$.
\item Using the recommender system under evaluation, compute predictions for the ratings of user $u$ for $i$ and for the  1000 random items from step 1.
\item Sort the 1001 items under consideration descendingly according to their predicted ratings. 
\item Denote by $p$ the position of item $i$ in that sorted list.
\item	Define a top-$N$ recommendation by selecting the first $N$ items from the sorted list. 
\item If $p \leq N$ then we have a \emph{hit}. Else we have a \emph{miss}. Thus, for each entry $(u,i,R_{ui}) \in T$ satisfying $R_{ui} = 5$ we either get a hit or a miss.
\end{enumerate}
Then, \emph{recall at $N$} is the average number of hits for $T$, i.e.,
\[
	\mathrm{recall}(N) := \frac{\text{number of hits}}{|T|}.
\] 
A closely related quantity is \emph{precision at $N$}. Precision specifies how much recall we have per item in the top-$N$ recommendation list, i.e., 
\[
	\mathrm{precision}(N) := \frac{\mathrm{recall}(N)}{N}.
\]

\section{Items from the long-tail}\label{sect:gergregkj}

Data available to train recommender systems usually follows a \emph{long-tail} distribution. I.e., a vast majority of the ratings available for training are ratings of a tiny fraction of all the items. For instance in the MovieLens 1M dataset, 5.5\% (i.e., 213 movies) of the most popular items amount for 33\% of all the ratings. As a user we might be a little disappointed by recommendations of very popular items as we may already be aware of those items. On the other hand, as a provider of the items, we want to push diversity in our product line. This motivated the testing methodology employed in~\cite{cremonesi2010performance} where the most popular items (6\%) are removed from the test set $T$.

\section{Proof of Theorem~\ref{Thm:NP.mindim}}\label{sect:proof.thm.NP.hardness}

The \emph{decision version} of \emph{MinDim} is \emph{NNM$_D$}.

\emph{NNM$_D$}. \emph{Problem instance}: $\bigl( \mathbb{P}_u[ \hat{E}_{i} = z ] \bigr)_{uiz \in \Omega}$ for some $\Omega \subseteq [U] \times [I] \times [Z]$ marking the probabilities that are known a priori. \emph{Acceptance condition}: we output \emph{yes} if and only if there exists a $D$-dimensional normalized nonnegative model (NNM) $\vec{p}_u$, $\vec{E}_{iz}$ such that $\vec{p}_u^T \vec{E}_{iz} = \mathbb{P}_u[ \hat{E}_{i} = z ]$ for all $(u,i,z) \in \Omega$.

Here we prove the following Theorem~\ref{Thm:NP}; it implies that \emph{MinDim} is \emph{NP}-hard because \emph{NNM$_D$} has to be accepted if and only if the minimizer of \emph{MinDim} is $\leq D$.

\begin{theorem}\label{Thm:NP}
	The decision problem \emph{NNM$_D$} is \emph{NP}-hard.
\end{theorem}

We prove Theorem~\ref{Thm:NP} in terms of a reduction from \emph{EXACT NMF$_k$} (see~\cite{vavasis2009complexity}) to \emph{NNM$_D$};

\emph{EXACT NMF$_k$} (see~\cite{vavasis2009complexity}). \emph{Problem instance}: a nonnegative matrix $M \in \AR^{m \times n}_+$ with $\rank(M) = k$. \emph{Acceptance condition}: we output \emph{yes} if and only if there exist nonnegative matrices $W \in \AR^{k \times m}_+$, $H \in \AR^{k \times n}_+$ with $k := \rank(M)$ such that $M = W^T H$.

A reduction from \emph{EXACT NMF$_k$} to \emph{NNM$_D$} suffices to prove the theorem because \emph{EXACT NMF$_k$} is \emph{NP}-hard;

\begin{theorem}[see \cite{vavasis2009complexity}]\label{Thm:NP.hardness.of.EXACTNMF}
	The decision problem \emph{EXACT NMF$_k$} is \emph{NP}-hard.
\end{theorem}

To prove theorem~\ref{Thm:NP} we show that there exists a polynomial time algorithm $\mathcal{A}$ with the two properties
	\[
		\mathcal{A}:  \{ \text{instances $M$ for \emph{EXACT NMF$_k$}} \} \rightarrow \{ \text{instances for \emph{NNM$_k$}} \}
	\]
	and
	\beq\label{fefwefeg}
		\mathcal{A}(M) \text{ \emph{yes} for NNM$_{k}$} \Leftrightarrow \text{\emph{yes} for \emph{EXACT NMF$_k$}} .
	\eeq
	The algorithm $\mathcal{A}$ we employ here does the following (recall that $M \in \AR^{m \times n}_+$). 
	\begin{itemize}
	\item 	Compute $M' \in \AR^{m \times n}_+$ by normalizing each row $M_{i,:}$ of $M$, i.e.,  
			\[
				M'_{i,:} = M_{i,:} / \Bigl( \sum_j M_{ij} \Bigr).
			\]
	\item		Set $U = m$, $I = 1$ $Z = n$ and
			\[
				\mathbb{P}_u[ \hat{E}_{1} = z ] = M'_{uz}.
			\]
	\item		Output $\bigl( \mathbb{P}_u[ \hat{E}_{1} = z ] \bigr)_{uiz \in \Omega}$ with $\Omega = [m] \times [1] \times [n]$.
	\end{itemize}
	We recognize that $\mathcal{A}$ indeed maps instances for \emph{EXACT NMF$_k$} to instances for \emph{NNM$_k$}. It is left to show~\eqref{fefwefeg}.
	
	``$\Rightarrow$": by assumption there exist $k$-dimensional distributions $\vec{p}_u$ and rating vectors $\vec{E}_{1z}$ such that $\vec{p}_u^T \vec{E}_{1z} = M'_{uz}$. Hence, the matrices $(\vec{p}_u, ..., \vec{p}_U)$ and $(\vec{E}_{11}, ..., \vec{E}_{1Z})$ realize the wanted $k$-dimensional NMF.
	
	``$\Leftarrow$": by assumption there exist $W \in \AR^{k \times m}_+$, $H \in \AR^{k \times n}_+$ such that $M = W^T H$. Therefore, setting
	\[
		W' := W \mathrm{diag}\Bigl(1/ \bigl( \sum_j M_{1j} \bigr), ..., 1/ \bigl( \sum_j M_{mj} \bigr)\Bigr)
	\]
	and $H' := H$, we get $M' = W'^T H'$. We denote by $W'_{:,u}$ and $H'_{:,z}$ columns of $W'$ and $H'$, respectively. Set $\vec{\eta} = \sum_z H'_{:,z}$ and $T = \mathrm{diag}(\vec{\eta})$. Then, $\vec{E}_{1z} := T^{-1}H'_{:,z}$ returns valid rating vectors in $\mathcal{E}'$. Moreover, the ansatz $\vec{p}_u := T W'_{:,u}$ yields valid probability distributions because
	\[
		\| \vec{p}_{u} \|_1 = \vec{p}_{u}^T (1,...,1)^T = \vec{p}^T \Bigl(  \sum_z \vec{E}_{1z}  \Bigr)
		 = \bigl( T W'_{:,u} \bigr)^T \Bigl(  \sum_z T^{-1} H'_{:,z}  \Bigr) = \sum_z M'_{uz} = 1.
	\]
	This proves the claim because $\vec{p}_{u}^T \vec{E}_{1z} = W_{:,u}'^T H'_{:,z} = M'_{uz} = \mathbb{P}_u[ \hat{E}_{1} = z ]$.

\section{Proof of Lemma~\ref{Lemma:fjekfjkejf}}\label{sect:proof.lemma.NMF.vs.NNM}

Set 
	\[
		M := \bigl( \mathbb{P}_u[ \hat{E}_{i} = z ] \bigr)_{u \in [U];i \in [I], z \in [Z]} \in \AR^{U \times IZ}.
	\]
	so that $M = A^T B$. By $D = \rank(A)$, there exist $u_1, ..., u_D$ such that the columns $\{ \vec{a}_{u_k} \}_{k=1}^D$ form a basis for $\AR^D$. By normalization of probability distributions,
	\[
		1 = \sum_{z} \mathbb{P}_{u_k}[ \hat{E}_{i} = z ] = \vec{a}_{u_k}^T \Bigl( \sum_{z} \vec{b}_{iz} \Bigr)
	\]
	for all $k \in [D]$. It follows that for all $i,i' \in [I]$
	\[
		\sum_{z} \vec{b}_{iz} = \sum_{z} \vec{b}_{i'z} =: \vec{\eta} \in \AR_+^D
	\]
	because$\{ \vec{a}_{u_k} \}_{k=1}^D$ is a basis. Assume that $\eta_j > 0$ for all $j$ (this will be proven afterwards). Then, $T := \mathrm{diag}(\vec{\eta})$ is invertible. Hence, $B' := T^{-1} B$ is well defined and the columns $\vec{b}'_{iz}$ of $B'$ satisfy
	\[
		\sum_z \vec{b}'_{iz} = T^{-1} \Bigl( \sum_z \vec{b}_{iz} \Bigr) = (1,...,1)^T
	\] 
	because $T^{-1} \vec{\eta} = (1,...,1)^T$. Consequently, the vectors $\vec{b}'_{iz}$ satisfy the normalization condition~\eqref{wdwdw3332}. Moreover, the nonnegative columns $\vec{a}'_{u}$ are valid probability distributions because
	\[
		\| \vec{a}'_{u} \|_1 = \vec{a}_{u}'^T (1,...,1)^T = \vec{a}_{u}'^T \Bigl(  \sum_z \vec{b}'_{iz}  \Bigr)\\ 
		 = \bigl( T \vec{a}_{u} \bigr)^T \Bigl(  \sum_z T^{-1} \vec{b}_{iz}  \Bigr) = \sum_z \mathbb{P}_u[ \hat{E}_{i} = z ] = 1.
	\]
	Therefore, the vectors $\vec{a}'_{u}$ and $\vec{b}'_{iz}$ constitute a valid NNM. This almost concludes the proof of the Lemma because
	\[
		\vec{a}_u'^T \vec{b}'_{iz} = \vec{a}_u^T \vec{b}_{iz} = \mathbb{P}_u[ \hat{E}_{i} = z ]
	\]
	for all $u,i,z$. It only remains to verify that $\eta_j \neq 0$ for all $j$. We provide an argument that is similar to an argument from~\cite{lee2014some} which was used to prove a relation between general positive semidefinite factorizations and quantum models. Assume that there exists $j$ such that $\eta_j = 0$. Thus,
	\beq\label{fewkjfekjs}
		0 = \eta_j = \Bigr( \frac{1}{I} \sum_{iz}  \vec{b}_{iz} \Bigr)_{j}.
	\eeq
	Now assume there exists $i,z$ with $\bigl( \vec{b}_{iz} \bigr)_{j} = \varepsilon > 0$. Then, $\sum_{iz}  ( \vec{b}_{iz} )_{j} \geq \varepsilon > 0$ because $( \vec{b}_{iz} )_{j} \geq 0$ for all $i,z$. This contradicts~\eqref{fewkjfekjs} and therefore, $\eta_j = 0$ implies $\bigl( \vec{b}_{iz} \bigr)_{j} = 0$ for all $i,z$. That, however, violates the condition $D = \rank(B)$ from the Lemma. We conclude that there cannot exist $j$ with $\eta_j = 0$.


\end{document}